\documentclass[conference]{IEEEtran}
\IEEEoverridecommandlockouts

\usepackage[skip=0pt]{caption}
\usepackage[skip=0pt]{subcaption}

\usepackage[top=0.75in, bottom=1.1in, left=0.75in, right=0.75in]{geometry}
\usepackage{latexsym}
\usepackage{amssymb}
\usepackage{amsmath}
\usepackage{amsthm}
\usepackage{booktabs}
\usepackage{enumitem}
\usepackage{graphicx}
\usepackage{color}
\usepackage{algorithm}
\usepackage{algpseudocode}
\usepackage{bm}

\usepackage{booktabs}
\usepackage{array} 
\usepackage{multirow}
\usepackage{makecell}

\usepackage{hyperref}
\newtheorem{definition}{Definition}
\usepackage{comment}
\usepackage{cite}
\usepackage{amsmath,amssymb,amsfonts}
\usepackage{graphicx}
\usepackage{textcomp}
\usepackage{xcolor}
\def\BibTeX{{\rm B\kern-.05em{\sc i\kern-.025em b}\kern-.08em
    T\kern-.1667em\lower.7ex\hbox{E}\kern-.125emX}}
\begin{document}

\title{Fairness-Constrained Optimization Attack in Federated Learning}

\author{
\IEEEauthorblockN{Harsh Kasyap\IEEEauthorrefmark{1}\IEEEauthorrefmark{2}, Minghong Fang\IEEEauthorrefmark{3}, Zhuqing Liu\IEEEauthorrefmark{4}, Carsten Maple\IEEEauthorrefmark{1}\IEEEauthorrefmark{5}, Somanath Tripathy\IEEEauthorrefmark{6}}
\IEEEauthorblockA{\IEEEauthorrefmark{1}The Alan Turing Institute London, Email: \{hkasyap, cm\}@turing.ac.uk}
\IEEEauthorblockA{\IEEEauthorrefmark{2}Indian Institute of Technology (BHU), Varanasi, Email: hkasyap.cse@iitbhu.ac.in}
\IEEEauthorblockA{\IEEEauthorrefmark{3}University of Louisville, USA, Email: minghong.fang@louisville.edu}
\IEEEauthorblockA{\IEEEauthorrefmark{4}University of North Texas, USA, Email: zhuqing.liu@unt.edu}
\IEEEauthorblockA{\IEEEauthorrefmark{5}University of Warwick, UK, Email: cm@warwick.ac.uk}
\IEEEauthorblockA{\IEEEauthorrefmark{6}Indian Institute of Technology Patna, Email: som@iitp.ac.in}
}
\iffalse
\author{Yalan Wang\inst{1}*
\and Bryan Kumara\inst{2}*
\and Harsh Kasyap\inst{2,3}
\and Liqun Chen\inst{1}
\and Sumanta Sarkar\inst{4}
\and Christopher J.P. Newton\inst{1}
\and Carsten Maple\inst{2,3}
% \and Christopher Newton\inst{1}
\and Ugur Ilker Atmaca\inst{2,3}}
\authorrunning{Y. Wang et al.}
\institute{University of Surrey, U.K. \and The Alan Turing Institute, London, U.K. \and The University of Warwick, U.K. \and The University of Essex, UK.
\thanks{Joint first authors.}
} 
\fi

\maketitle

\begin{abstract}
Federated learning (FL) is a privacy-preserving machine learning technique that facilitates collaboration among participants across demographics. FL enables model sharing, while restricting the movement of data. Since FL provides participants with independence over their training data, it becomes susceptible to poisoning attacks. Such collaboration also propagates bias among the participants, even unintentionally, due to different data distribution or historical bias present in the data. This paper proposes an intentional fairness attack, where a client maliciously sends a biased model, by increasing the fairness loss while training, even considering homogeneous data distribution. The fairness loss is calculated by solving an optimization problem for fairness metrics such as demographic parity and equalized odds. The attack is insidious and hard to detect, as it maintains global accuracy even after increasing the bias. We evaluate our attack against the state-of-the-art Byzantine-robust and fairness-aware aggregation schemes over different datasets, in various settings. The empirical results demonstrate the attack efficacy by increasing the bias up to 90\%, even in the presence of a single malicious client in the FL system.
\end{abstract}

\section{Introduction}
Federated learning (FL) has evolved as a promising collaborative learning framework~\cite{mcmahan2016federated}, and has been used in critical applications such as healthcare~\cite{kulkarni2021dnet}. Unlike traditional centralized machine learning (ML) systems, FL enables decentralized data training across multiple devices, preserving privacy by only allowing the movement of the ML model.
FL, due to its distributed nature, has been extensively studied in the context of poisoning attacks~\cite{fang2020local,kasyap2022hidden}.
New challenges around fairness and bias have also been discussed in FL~\cite{chang2023bias}. Sensitive personal attributes, such as race, gender, and socioeconomic status, are embedded within decentralized datasets, increasing the risk of perpetuating and even amplifying biases. 

FL’s inherent design inadvertently allows participants with overrepresented or underrepresented groups to propagate bias to the global model, causing fairness issues, as unequal representation across participants may skew the learning process. Moreover, adversaries can intentionally frame a poisoned model by conducting fairness attacks, subtly manipulating local data or models to induce biased outcomes in the global model. For example, it may result in higher risk assessments in finance and affect access to loans and interest rates for certain racial groups. Bias has been reported in ML models of Amazon's recruitment system\footnote{https://www.bbc.co.uk/news/technology-45809919} favoring male candidates and COMPAS system predicting biased recidivism~\cite{dressel2018accuracy}. Since the fairness issue has been highly concerning even when introduced unintentionally, it should be studied when malicious participants intentionally introduce bias in the system.

Existing poisoning attacks manipulate the local data or model, aiming to degrade the global model accuracy indiscriminately of the test samples~\cite{fang2020local}. There can also be targeted poisoning attacks, which degrade the performance of a specific class and become difficult to detect~\cite{bagdasaryan2020backdoor}. However, such attacks cause predictive loss, rather than bias in general. There have been a few attempts in the literature~\cite{ijcai2024p51,rance2024attacksfairnessfederatedlearning}, investigating intentional fairness attacks in federated learning. Authors in~\cite{rance2024attacksfairnessfederatedlearning} framed a biased model by training on attribute-specific data only. However, data manipulation attacks are easier to detect since they also exhibit poor predictive performance. Authors in~\cite{ijcai2024p51} proposed a model poisoning-based fairness attack, which injects biases into the redundant space of the model, by solving an optimization problem. Such that it does not impact the overall predictive performance and only causes fairness loss. However, it suffers noticeable drops in global accuracy and is costly in terms of finding both the redundant space in the model and solving the optimization problem.

Existing Byzantine-robust aggregation schemes~\cite{blanchard2017machine,yin2018byzantine,fang2025we,fang2024byzantine} are primarily designed to defend against arbitrary or adversarial updates from a subset of participants in federated learning. These methods aim to maintain the integrity and utility of the global model by filtering or down-weighting malicious or outlier updates that could degrade model accuracy. Similarly, fairness-aware aggregation schemes such as FairFed~\cite{ezzeldin2023fairfed} and FairTrade~\cite{badar2024fairtrade} have been proposed to address naturally occurring disparities in model performance across sensitive subgroups. These approaches introduce additional fairness constraints or regularization terms to guide the learning process toward equitable outcomes. However, these defenses typically assume that any unfairness or bias arises organically from skewed or imbalanced data distributions, rather than from deliberate adversarial behavior. This gap in the threat model opens a critical vulnerability: the potential for an intentional adversary to strategically poison local models not to degrade accuracy, but to amplify unfairness in the global model—thereby violating fairness constraints without necessarily triggering traditional Byzantine defenses. Motivated by this observation, we design a targeted model poisoning attack that specifically manipulates fairness loss.

\noindent\textbf{Our work.} This paper proposes an intentional fairness attack in FL, which aims to increase bias in the global model without a noticeable decrease in the accuracy of the classification task. It requires the malicious participants to solve fairness constrained optimization for respective fairness metrics, such as demographic parity or equalized odds. Adding the fairness loss while optimizing the classification task maintains the prediction accuracy and successfully induces bias in the global model. Building on the reduction-based framework for fair classification proposed by Agarwal et al.~\cite{agarwal2018reductions}, we investigate the susceptibility of such methods to adversarial influence. In this framework, the task of learning a fair classifier is reduced to solving a sequence of cost-sensitive classification problems, allowing standard cost-sensitive learners to be used as oracles within a convex optimization procedure. We design an attack that seeks to increase the fairness loss by strategically perturbing the inputs to the cost-sensitive classification subproblems, thereby impacting the satisfaction of fairness constraints in the final classifier without substantially degrading predictive performance. 

%The proposed attack maintains the properties of (a) high model utility, (b) attribute-specific biases and (c) effective against Byzantine-robust and fairness-aware aggregation schemes. To evaluate the effectiveness and practicality of the attack, we consider the number of malicious participants to be less; even a single participant is able to induce and increase bias up to 90\%. 

\section{Background and Related Work}

\subsection{Federated Learning (FL)}

FL is a collaborative machine learning paradigm, emphasising privacy preservation ~\cite{mcmahan2016federated}. In the system, we assume a central server ($S$) and $n$ participants ($p_1, p_2, \ldots, p_n$) with local private training dataset $d_{i}$ (for each participant $i \in n$). The central server $\mathcal{S}$ sends the global model $w^{t}$ to each participant for local model training in each communication round $t$. Then, each participant performs local model training as follows:
\begin{equation}
\label{eq:fltrain}
w^{t+1}_{i} \leftarrow w_{i}^t - \eta\Delta l(w_{i}^t;b \in d_{i}),
\end{equation}
where $\eta$ is the local learning rate and $b$ is a batch from local participant dataset $d_i$, $l$ is the local training objective of client. 
After local model training, each participant $p_i$ sends its local model update $\Delta w_{i}$ to $\mathcal{S}$. $\mathcal{S}$ computes the global model $w^{t+1}$ for the next iteration, using aggregation rules such as FedAvg~\cite{mcmahan2017communication}. It performs a weighted averaging of the local model updates as the following:
\begin{equation}
\label{eq:fedavg}
w^{t+1} \leftarrow w^{t} + \sum_{i=1}^{n} \frac{k_i}{ \sum_{j=1}^{n}k_{j}}\Delta w_{i}^{t+1},
\end{equation}
where $k_{i}$ is the number of samples held by $p_i$. This process repeats until convergence. However, the global model can sometimes exhibit demographic biases due to the heterogeneity of the datasets across participants~\cite{ezzeldin2023fairfed}.

\subsection{Fairness Metrics}

\noindent\textbf{Group Fairness.} Group fairness in FL ensures that the model's predictions or outcomes are not biased against specific groups defined by any sensitive attribute $A$. $A$ represents a demographic characteristic such as gender, race, age, or any other feature that could introduce bias into the model's behavior. The goal of group fairness is to guarantee that the model treats each group (\textit{i.e.}, different values of $A$) equitably, such that the model does not disproportionately favor one group over another. This can be measured using different fairness metrics, such as demographic parity and equalized odds.

\begin{definition}[Demographic Parity (DP)]\label{def-dp}
Demographic parity aims to ensure equal probability of a positive outcome ($\mathrm{\hat{Y}} = 1$) across different groups ($A = 0 \text{ and } 1$) of a given sensitive attribute $A$~\cite{dwork2012fairness}. Formally, it is defined as,
\begin{align}
\label{dp_def}
M_{\text{DP}} = 
\left|\mathrm{P}({\mathrm{\hat{Y}} = 1} \mid \mathrm{A}=0) - \mathrm{P}({\mathrm{\hat{Y}} = 1} \mid \mathrm{A}=1)\right| \leq \epsilon.  
\end{align}
\end{definition}
A smaller value of $\epsilon$ signifies a higher degree of fairness. It ensures that the model does not favour one group over another in terms of the likelihood of positive outcomes. 

\begin{definition}[Equalized Odds (EOD)]\label{def-eo}
%Refer our paper.
Equalized odds aims to ensure fairness in the true positive rates (TPR) and false positive rates (FPR), across different groups~\cite{kusner2017counterfactual}. Formally, it is defined as,
\begin{align}
\label{eo_def}
M_{\text{EOD}} = 
&\left| P(\hat{Y}=1 \mid A=0, Y=y)  - \right. \nonumber \\
&\qquad \left. P(\hat{Y}=1 \mid A=1, Y=y) \right| \leq \epsilon.
\end{align}
\end{definition}
A smaller value of $\epsilon$ indicates a higher degree of fairness for the model. Equalized odds is satisfied when $\epsilon = 0$. 
Unfortunately, this can be very hard to achieve in practice, so it makes sense to relax the EOD criterion and consider a modified version of the EOD equation with \textit{y=1} for equalizing TPR (equal opportunity), or \textit{y=0} for equalizing FPR. Considering TPR in a relaxed setting is called Equal opportunity, which ensures that the model predicts an equal chance for members of all groups who qualify for a positive outcome.
%\smallskip

\subsection{Related Work}

\noindent\textbf{Fairness Attacks.} 
Local model poisoning attacks in FL have been extensively studied as a form of adversarial behavior where malicious clients attempt to compromise the integrity of the global model by manipulating their locally trained models before uploading them to the central server~\cite{fang2020local,ijcai2024p51}. These attacks typically aim to degrade the overall performance of the global model, often without being detected, by subtly embedding harmful updates into the federated aggregation process.

Among these, EAB-FL~\cite{ijcai2024p51} represents a novel class of local model poisoning attacks that shifts the focus from reducing overall accuracy to exacerbating group-level disparities in model performance. %Instead of causing a significant drop in the model’s general utility, which might raise suspicion, 
EAB-FL is carefully crafted to maintain the global model accuracy while selectively harming the model’s performance on specific subpopulations or sensitive groups, thereby increasing group unfairness. The attack proceeds in two main stages. First, it employs Layer-wise Relevance Propagation (LRP), to analyze the locally trained model and identify redundant parameter spaces, which are regions of the model that do not significantly contribute to the overall prediction performance. These redundant spaces are exploited as stealthy channels through which adversarial modifications can be injected without impacting the global model's utility. Next, EAB-FL formulates and solves an optimization problem over a strategically chosen subset of the attacker’s local dataset. The goal of this optimization is to craft model updates that, when aggregated into the global model, disproportionately degrade the performance for a targeted group (e.g., a demographic defined by race, gender, or age). 
%\smallskip

\noindent\textbf{Fairness-aware Aggregation.} %In Federated Learning (FL), ensuring fairness across different clients or user groups has emerged as a critical concern, especially when data is distributed non-uniformly across participants. To address these challenges, several aggregation strategies have been proposed that explicitly incorporate fairness considerations during the model aggregation phase.  
FairFed~\cite{ezzeldin2023fairfed} is one such method that introduces fairness-aware aggregation by computing a fairness deviation score for each client. Specifically, it measures the difference between a client's local fairness metric (e.g., disparity in model performance across subgroups) and the global fairness objective. Clients whose local models contribute more positively toward improving global fairness are assigned higher aggregation weights, while those that exacerbate unfairness are down-weighted. This dynamic weighting mechanism enables FairFed to perform well in highly heterogeneous settings. However, in homogeneous settings, where client data distributions are similar, then FairFed tends to assign nearly identical fairness scores to all clients. FairTrade~\cite{badar2024fairtrade} empirically shows that existing schemes achieve fairness at the cost of balanced accuracy. They applied fairness-constrained optimization locally and further utilized multiobjective optimization to find a balance between balanced accuracy and fairness. Other poisoning-resistant aggregation schemes such as Krum, Multi-Krum~\cite{blanchard2017machine}, Median, and Trimmed-Mean~\cite{yin2018byzantine} are designed to mitigate poisoning attacks by performing statistical operations such as Euclidean distance or cosine similarity.
%%\smallskip

\begin{comment}

\section{Threat Model}
%
We assume that the adversary has taken control of a small subset of participants with the intent to manipulate the local model training process. These compromised participants are referred to as malicious clients.
%
The server also deploys poisoning-resistant and fairness-aware aggregation schemes to detect such participants.
%\smallskip

\noindent\textbf{Adversary's goal.} The adversary does not intend to degrade the overall model's utility, but rather aims to increase the algorithmic bias (in terms of metrics such as demographic parity and equalised oddds) in the global model while maintaining global accuracy.
%\smallskip

\noindent\textbf{Adversary's capability.} The adversary can corrupt the local model training process on the malicious clients. However, they do not manipulate the local training dataset, rather they can deploy an optimization based approach to induce bias for a targeted group.
%\smallskip

\noindent\textbf{Adversary's knowledge.} 
%
The adversary gains access to the structure and parameters of the global model in each round, as the malicious clients receive the global model from the central server during every round. The adversary has no information about the local models of the honest participants and the aggregation rule deployed on the server.
\end{comment}
%%%%%%%%%%%%%%%%%%%%%%%%%%%

\section{Our Attack}
Fairness in FL has emerged as a pressing concern due to the decentralized nature of data and the inherent biases present in local client datasets. While existing methods primarily focus on enhancing fairness during the training process, adversarial entities can exploit fairness constraints to undermine the learning process or distort the model's performance across specific subgroups. This necessitates the development of a fairness attack strategy designed to compromise the fairness of the global model systematically, while maintaining plausible deniability by embedding the attack within the standard local optimization framework. Our proposed approach investigates how malicious clients can manipulate fairness metrics during local updates, ultimately diminishing the equity of the global model.

The core concept of our fairness attack lies in introducing a malicious objective that deliberately alters local updates to degrade the global model's fairness. By modifying the training objectives of malicious clients, the attack generates updates that appear legitimate but amplify disparities in fairness metrics, such as demographic parity or equal opportunity. This exploitation leverages the FL setup, where updates are aggregated without access to raw client data, making it challenging to detect fairness violations at their source.

Formally, let \( l(w) \) represent the original local training objective of a client before the attack. For a local classifier parameterized by \( w \), the optimization problem at one malicious client side is redefined as:  
\begin{align}
\label{new_obj}
\min_{w} l(w) - \lambda \cdot M_*,
\end{align}
where \( M_* \) denotes a fairness metric, such as demographic parity (defined in Eq.~(\ref{dp_def})) or equalized odds (defined in Eq.~(\ref{eo_def})), and \( \lambda \) is a regularization parameter that controls the balance between the standard training objective and the malicious fairness manipulation.
In many cases, \( M_* \) is non-differentiable, which complicates the use of gradient-based optimization methods. To address this, we can approximate the gradient of \( M_* \) using the finite difference method. Specifically, by slightly perturbing \( w \) and evaluating \( M_* \) at these perturbed points, we can numerically estimate the gradient. This approach approximates \( \nabla_w M_* \) as follows:  
\begin{align}
\nabla_w M_* \approx \frac{M_*(w + \epsilon) - M_*(w - \epsilon)}{2\epsilon},
\end{align}
where \( \epsilon \) represents a small perturbation. This method enables effective optimization even when \( M_* \) lacks differentiability.
%\smallskip

\noindent\textbf{DP-based fairness-constrained optimization.} 
The demographic parity (DP) constraint requires that the prediction is independent of the sensitive attribute. In a binary classification setting with a binary sensitive attribute, this can be formalized by ensuring that the conditional probabilities of positive predictions are approximately equal across the sensitive groups. Specifically, the DP constraint is encoded through a set of linear constraints represented by the matrix \textit{Q} and the vector of conditional moments $\eta(f)$. 

To formalize the calculation of \( M_* \), for DP with a given budget $\epsilon$, a set of linear constraints is generated in the form of $Q\eta(f)\leq \epsilon$. %, 
It is based on how the prediction $f(X)$ varies for different subsets of the data,
considering the input data $(X)$, true labels $(Y)$, and a sensitive attribute $(A)$. The construction of \textit{Q} ensures that for each group defined by the sensitive attribute \textit{A}, the difference in conditional expectations of the model’s predictions is captured.

Assuming a binary classification task and a binary sensitive attribute, DP can be expressed as a set of two equality constraints. The elements of $Q$ are initialized to form a set of linear constraints as below.

\begin{equation}
Q_{\left(A, \Delta^{+}\right), A^{\prime}}=\left\{\begin{array}{cl}
1 & \text { if } A^{\prime}=A \\
-1 & \text { otherwise. }
\end{array}\right.
\end{equation}

\begin{equation}
Q_{\left(A, \Delta^{-}\right), A^{\prime}}=\left\{\begin{array}{cl}
-1 & \text { if } A^{\prime}=A \\
1 & \text { otherwise. }
\end{array}\right.
\end{equation}

Each equality constraint can be formulated as a pair of positive $\left(\Delta^{+}:=\eta_A(f)-\eta_{\{X \backslash A\}}(f) \leq 0\right)$ and negative $\left(\Delta^{-}:=-\eta_A(f)+\eta_{\{X \backslash A\}}(f) \leq 0\right)$ inequality constraints. $Q_{\left(A, \Delta^{+}\right), A^{\prime}}$ assign a value of 1 when $A^{\prime} = A$ and -1 otherwise, capturing the deviation between the group \textit{A} and the rest of the population for the positive constraint $(\Delta^{+})$. $\Delta^{+}$ ensures that the conditional moment for group \textit{A} minus that for the rest should not be too large (no large positive deviation). Similarly, $Q_{\left(A, \Delta^{-}\right), A^{\prime}}$ captures the deviation for the negative constraint $(\Delta^{-})$, flipping the signs accordingly. $\Delta^{-}$ ensures that the conditional moment for the rest minus group \textit{A} should also not be too large (no large negative deviation).

When computing the fairness loss, a greater emphasis on larger errors is placed by using the L2-norm of the constraint, and the fairness metric is re-written as:

\begin{equation}
M_*=\lambda \times \|(\operatorname{\text{ReLU}}(Q \eta(f)))-\epsilon\|_2.
\end{equation}

$M_{*}$ is proportional (via multiplier $\lambda$) to the L2-
norm of the positive part (via ReLU) of the constraint violations $(Q \eta(f)))-\epsilon$. The use of the ReLU ensures that only violations exceeding the fairness budget are penalized, and the L2-norm aggregates the magnitude of these violations, placing more emphasis on larger violations. Similar constructions can be extended to other fairness definitions, such as equalized odds, by appropriately modifying how \textit{Q} and $\eta(f)$ are computed.
%\smallskip

The fairness constraint framework naturally extends to settings with more than two groups defined by the sensitive attribute $(A)$. The goal remains to ensure that the conditional expectation of predictions is approximately equal across all groups. Let $A \in\left\{a_1, a_2, \ldots, a_k\right\}$ denote the $k$ distinct sensitive groups. To enforce fairness among all pairs of groups, we generate $\binom{k}{2}$ pairs of linear inequality constraints-each pair consisting of a positive and a negative version of the moment difference. For every pair $\left(a_i, a_j\right)$, the positive constraint $\eta_{a_i}(f)-\eta_{a_j}(f) \leq \epsilon$ ensures that group $a_i$ is not unduly favored over $a_j$, while the corresponding negative constraint $-\eta_{a_i}(f)+\eta_{a_j}(f) \leq \epsilon$ guarantees symmetry. 

In the proposed attack framework, the malicious clients incorporate \( M_* \) to quantify the model's adherence to fairness criteria, such as equalizing predictions across demographic groups (demographic parity) or ensuring consistent true positive rates across groups (equal opportunity). The introduction of a negative scaling term \( -\lambda \cdot M_* \) enables the client to systematically direct local model updates in a manner that degrades fairness. 

\begin{table}[t]
% \tiny
\caption{Fairness attack evaluation on the Adult dataset.}
\label{tab:adult}
\centering
%\footnotesize
\scriptsize
%\tiny
\addtolength{\tabcolsep}{-2.66pt}
\begin{tabular}{|l|r|r|r|r|r|r|r|r|r|}
\hline
    & \multicolumn{3}{|c|}{No attack}& \multicolumn{3}{|c|}{One malicious client}& \multicolumn{3}{|c|}{Two malicious clients}\\ \hline
Scheme    & \multicolumn{1}{|l|}{Acc}       & \multicolumn{1}{|l|}{DP}    & \multicolumn{1}{|l|}{EOD} & \multicolumn{1}{|l|}{Acc}       & \multicolumn{1}{|l|}{DP}    & \multicolumn{1}{|l|}{EOD} & \multicolumn{1}{|l|}{Acc}       & \multicolumn{1}{|l|}{DP}    & \multicolumn{1}{|l|}{EOD} \\ \hline
FedAvg    & \multicolumn{1}{|r|}{0.708}  & \multicolumn{1}{|r|}{0.371} & 0.298                    & \multicolumn{1}{|r|}{0.724}  & \multicolumn{1}{|r|}{0.443} & 0.369                    & \multicolumn{1}{|r|}{0.713}  & \multicolumn{1}{|r|}{0.498} & 0.430                     \\ \hline
Krum      & \multicolumn{1}{|r|}{0.670}   & \multicolumn{1}{|r|}{0.267} & 0.187                    & \multicolumn{1}{|r|}{0.726}  & \multicolumn{1}{|r|}{0.383} & 0.298                    & \multicolumn{1}{|r|}{0.717}  & \multicolumn{1}{|r|}{0.389} & 0.306                    \\ \hline
MKrum     & \multicolumn{1}{|r|}{0.696}  & \multicolumn{1}{|r|}{0.353} & 0.280                     & \multicolumn{1}{|r|}{0.717}  & \multicolumn{1}{|r|}{0.384} & 0.309                    & \multicolumn{1}{|r|}{0.724}  & \multicolumn{1}{|r|}{0.390}  & 0.318                    \\ \hline
Median    & \multicolumn{1}{|r|}{0.724}  & \multicolumn{1}{|r|}{0.387} & 0.314                    & \multicolumn{1}{|r|}{0.720}   & \multicolumn{1}{|r|}{0.408} & 0.339                    & \multicolumn{1}{|r|}{0.721}  & \multicolumn{1}{|r|}{0.425} & 0.354                    \\ \hline
TMean     & \multicolumn{1}{|r|}{0.717}  & \multicolumn{1}{|r|}{0.375} & 0.302                    & \multicolumn{1}{|r|}{0.719}  & \multicolumn{1}{|r|}{0.414} & 0.342                    & \multicolumn{1}{|r|}{0.717}  & \multicolumn{1}{|r|}{0.432} & 0.363                    \\ \hline
FairFed & \multicolumn{1}{|r|}{0.708}  & \multicolumn{1}{|r|}{0.368}  & 0.296                    & \multicolumn{1}{|r|}{0.723}    & \multicolumn{1}{|r|}{0.442} & 0.370                    & \multicolumn{1}{|r|}{0.714}   & \multicolumn{1}{|r|}{0.498} & 0.428                    \\ \hline
FairTrade & \multicolumn{1}{|r|}{0.708}  & \multicolumn{1}{|r|}{0.020}  & 0.088                    & \multicolumn{1}{|r|}{0.700}    & \multicolumn{1}{|r|}{0.049} & 0.091                    & \multicolumn{1}{|r|}{0.697}   & \multicolumn{1}{|r|}{0.084} & 0.096                    \\ \hline
\end{tabular}
\end{table}

\begin{table}[t]
\caption{Fairness attack evaluation [attribute-based distribution]. Note that a noticeable decrease in accuracy is observed only here due to a high class imbalance.}
\label{tab:adult-attr}
\centering
\scriptsize
%\tiny
\addtolength{\tabcolsep}{-2.66pt}
\begin{tabular}{|l|rrr|rrr|rrr|}
\hline
  & \multicolumn{3}{c|}{No attack}                                                          & \multicolumn{3}{c|}{One malicious client}                                                           & \multicolumn{3}{c|}{Two malicious clients}                                                           \\ \hline
Scheme    & \multicolumn{1}{l|}{Acc}   & \multicolumn{1}{l|}{DP}    & \multicolumn{1}{l|}{EOD} & \multicolumn{1}{l|}{Acc}   & \multicolumn{1}{l|}{DP}    & \multicolumn{1}{l|}{EOD} & \multicolumn{1}{l|}{Acc}   & \multicolumn{1}{l|}{DP}    & \multicolumn{1}{l|}{EOD} \\ \hline
FedAvg    & \multicolumn{1}{r|}{0.723} & \multicolumn{1}{r|}{0.314} & 0.224                    & \multicolumn{1}{r|}{0.596} & \multicolumn{1}{r|}{0.591} & 0.577                    & \multicolumn{1}{r|}{0.503} & \multicolumn{1}{r|}{0.986} & 0.997                    \\ \hline
Krum      & \multicolumn{1}{r|}{0.746} & \multicolumn{1}{r|}{0.328} & 0.253                    & \multicolumn{1}{r|}{0.644} & \multicolumn{1}{r|}{0.403} & 0.365                    & \multicolumn{1}{r|}{0.482} & \multicolumn{1}{r|}{0.987} & 0.954                    \\ \hline
MKrum     & \multicolumn{1}{r|}{0.750} & \multicolumn{1}{r|}{0.329} & 0.254                    & \multicolumn{1}{r|}{0.710} & \multicolumn{1}{r|}{0.371} & 0.301                    & \multicolumn{1}{r|}{0.501} & \multicolumn{1}{r|}{0.983} & 0.947                    \\ \hline
Median    & \multicolumn{1}{r|}{0.759} & \multicolumn{1}{r|}{0.336} & 0.256                    & \multicolumn{1}{r|}{0.677} & \multicolumn{1}{r|}{0.415} & 0.364                    & \multicolumn{1}{r|}{0.491} & \multicolumn{1}{r|}{0.946} & 0.976                    \\ \hline
TMean     & \multicolumn{1}{r|}{0.764} & \multicolumn{1}{r|}{0.339} & 0.256                    & \multicolumn{1}{r|}{0.678} & \multicolumn{1}{r|}{0.420} & 0.368                    & \multicolumn{1}{r|}{0.510} & \multicolumn{1}{r|}{0.976} & 0.983                    \\ \hline
FairFed   & \multicolumn{1}{r|}{0.708} & \multicolumn{1}{r|}{0.324} & 0.240                    & \multicolumn{1}{r|}{0.644} & \multicolumn{1}{r|}{0.605} & 0.576                    & \multicolumn{1}{r|}{0.515} & \multicolumn{1}{r|}{0.983} & 0.993                    \\ \hline
FairTrade & \multicolumn{1}{r|}{0.753} & \multicolumn{1}{r|}{0.045} & 0.079                    & \multicolumn{1}{r|}{0.703} & \multicolumn{1}{r|}{0.067} & 0.087                    & \multicolumn{1}{r|}{0.578} & \multicolumn{1}{r|}{0.085} & 0.096                    \\ \hline
\end{tabular}
\end{table}

\begin{table}[t]
\caption{Fairness attack comparison, with one malicious client. Bold represents the best result in each column - highest Accuracy, DP and EOD.}
\label{tab:comp1}
\centering
\scriptsize
%\tiny
\addtolength{\tabcolsep}{-2.66pt}
%\resizebox{\textwidth}{!}{
\begin{tabular}{|l|r|r|r|r|r|r|r|r|r|}
\hline
 & \multicolumn{3}{|c|}{DYN-OPT}& \multicolumn{3}{|c|}{EAB}& \multicolumn{3}{|c|}{Our attack}\\ \hline
Scheme                     & \multicolumn{1}{|l|}{Acc}       & \multicolumn{1}{|l|}{DP}    & \multicolumn{1}{|l|}{EOD} & \multicolumn{1}{|l|}{Acc}       & \multicolumn{1}{|l|}{DP}    & \multicolumn{1}{|l|}{EOD} & \multicolumn{1}{|l|}{Acc}       & \multicolumn{1}{|l|}{DP}    & \multicolumn{1}{|l|}{EOD}             \\ \hline
FedAvg                     & \multicolumn{1}{|r|}{0.683}   & \multicolumn{1}{|r|}{0.365} & 0.298& \multicolumn{1}{|r|}{0.701}  & \multicolumn{1}{|r|}{0.371} & 0.297                    & \multicolumn{1}{|r|}{\textbf{0.724}}    & \multicolumn{1}{|r|}{\textbf{0.443}} & \textbf{0.369}                    \\ \hline
Krum                       & \multicolumn{1}{|r|}{\textbf{0.727}}  & \multicolumn{1}{|r|}{\textbf{0.408}} & \textbf{0.335}& \multicolumn{1}{|r|}{0.717}  & \multicolumn{1}{|r|}{0.379} & 0.291                    & \multicolumn{1}{|r|}{0.726}  & \multicolumn{1}{|r|}{0.383} & 0.298                    \\ \hline
MKrum                      & \multicolumn{1}{|r|}{\textbf{0.746}}  & \multicolumn{1}{|r|}{0.383}  & 0.304& \multicolumn{1}{|r|}{0.739}  & \multicolumn{1}{|r|}{0.380} & 0.302                    & \multicolumn{1}{|r|}{0.717}  & \multicolumn{1}{|r|}{\textbf{0.384}} & \textbf{0.309}                    \\ \hline
Median                     & \multicolumn{1}{|r|}{\textbf{0.747}}  & \multicolumn{1}{|r|}{0.395} & 0.315& \multicolumn{1}{|r|}{0.744}   & \multicolumn{1}{|r|}{0.389} & 0.313                    & \multicolumn{1}{|r|}{0.720}  & \multicolumn{1}{|r|}{\textbf{0.408}} & \textbf{0.339}                    \\ \hline
TMean                      & \multicolumn{1}{|r|}{\textbf{0.755}}  & \multicolumn{1}{|r|}{0.403} & 0.321& \multicolumn{1}{|r|}{0.748}  & \multicolumn{1}{|r|}{0.391} & 0.310                     & \multicolumn{1}{|r|}{0.719}  & \multicolumn{1}{|r|}{\textbf{0.414}} & \textbf{0.342}                    \\ \hline
FairFed                  & \multicolumn{1}{|r|}{0.695}  & \multicolumn{1}{|r|}{0.378} & 0.304& \multicolumn{1}{|r|}{0.711}    & \multicolumn{1}{|r|}{0.391} & 0.310                    & \multicolumn{1}{|r|}{\textbf{0.723}}   & \multicolumn{1}{|r|}{\textbf{0.442}} & \textbf{0.370}                    \\ \hline
FairTrade                  & \multicolumn{1}{|r|}{0.687}  & \multicolumn{1}{|r|}{0.031} & 0.062& \multicolumn{1}{|r|}{0.674}    & \multicolumn{1}{|r|}{0.033} & 0.072                    & \multicolumn{1}{|r|}{\textbf{0.700}}   & \multicolumn{1}{|r|}{\textbf{0.049}} & \textbf{0.091}                    \\ \hline
%\\
%\hline
\end{tabular}%}
\end{table}

The regularization parameter \( \lambda \) is pivotal in determining the attack's intensity. A higher \( \lambda \) prioritizes fairness manipulation over the original objective \( l(w) \), amplifying the attack's impact, while a lower \( \lambda \) balances fairness manipulation with standard optimization, ensuring the attack remains stealthy. This formulation empowers malicious clients to produce harmful updates that compromise the fairness of the aggregated global model without noticeably deviating from standard optimization patterns, thereby evading detection.
Through this optimization mechanism, malicious clients exploit the inherent trust in FL systems to subtly and systematically disrupt fairness, posing a significant challenge to existing defense mechanisms.

%%%%%%%%%%%%%%%%%%%%%%%%%%%
\section{Experimental Evaluation}

\subsection{Experimental Setup}

\textbf{Datasets.} The proposed attack has been evaluated on five datasets: Adult~\cite{bache2013uci}, Bank~\cite{bache2013uci}, Default~\cite{bache2013uci}, Law~\cite{wightman1998lsac}, and KDD~\cite{bache2013uci}, which encompasses various configurations of attributes, instances, sensitive features, and class imbalance levels. The ratios of positive to negative classes in these datasets are as follows: 1:3.0 for Adult, 1:7.87 for Bank, 1:3.52 for Default, 1:3.50 for Law, and 1:15.11 for KDD. These datasets have been well used for fairness studies. The datasets are partitioned either randomly or using specific attributes (to ensure the presence of non-iid data) to simulate realistic scenarios, similar to~\cite{badar2024fairtrade}. 

\noindent\textbf{Evaluated Defenses.} FedAvg~\cite{mcmahan2016federated}, Krum and Multi-Krum~\cite{blanchard2017machine}, Median and Trimmed Mean~\cite{yin2018byzantine}, FairFed~\cite{ezzeldin2023fairfed} and FairTrade~\cite{badar2024fairtrade} are considered different poisoning-resistant and fairness-aware aggregation rules deployed on the server. %It is important to note that, FairFed and FairTrade are employed in coordination with participants and the server, where the honest participants engage in prior communication (privately) to share their bias statistics with the server.
%\smallskip

\noindent\textbf{Compared attacks.} The proposed attack is compared with optimization-based local model poisoning attack DYN-OPT~\cite{shejwalkar2021manipulating} and fairness attack EAB-FL~\cite{ijcai2024p51}.
%\smallskip

\noindent\textbf{Evaluation metrics.} We consider Accuracy (Acc), Demographic parity (DP), and Equalised Odds (EOD) as evaluation metrics. Our proposed attack aims to increase DP and EOD, while maintaining accuracy as the no-attack scenario.
%\smallskip

\noindent\textbf{Parameter settings.} We employ a feedforward neural network comprising three fully connected layers with hidden dimensions of 64 and 32, respectively. Each hidden layer is followed by a ReLU activation, and a final sigmoid activation is applied to produce a scalar output suitable for binary classification. The model architecture is lightweight, which is suitable for training over these datasets. Unless specified, we consider 10 participants, 50 communication rounds, and 1 local epoch per round. Malicious participants are considered as 10\% (1) and 20\% (2) in each round. We also evaluate the efficacy of the attack with 50 participants, considering 1 and 2 malicious participants. The learning rate is set as 0.001, and the regularisation constraint is 100. Unless specified, we utilize DP-based fairness-constrained optimization for framing the attack, and the data distribution is considered random. We ran the experiments 10 times each and reported the average result.

\begin{table}[t]
\caption{Fairness attack evaluation on the Law dataset.}
\label{tab:law}
\centering
\scriptsize
%\tiny
\addtolength{\tabcolsep}{-2.66pt}
\begin{tabular}{|l|r|r|r|r|r|r|r|r|r|}
\hline
 & \multicolumn{3}{|c|}{No attack}& \multicolumn{3}{|c|}{One malicious client}& \multicolumn{3}{|c|}{Two malicious clients}\\ \hline
Scheme    & \multicolumn{1}{|l|}{Acc}     & \multicolumn{1}{|l|}{DP} & \multicolumn{1}{|l|}{EOD} & \multicolumn{1}{|l|}{Acc}     & \multicolumn{1}{|l|}{DP} & \multicolumn{1}{|l|}{EOD} & \multicolumn{1}{|l|}{Acc}     & \multicolumn{1}{|l|}{DP} & \multicolumn{1}{|l|}{EOD} \\ \hline
FedAvg    & 0.832                    & 0.003                   & 0.021                    & 0.824                    & 0.042                   & 0.062                    & 0.823                    & 0.077                   & 0.126                    \\ \hline
Krum      & 0.851                    & 0.021                   & 0.065                    & 0.840                     & 0.026                   & 0.061                    & 0.794                    & 0.019                   & 0.037                    \\ \hline
MKrum     & 0.820                     & 0.010                    & 0.029                    & 0.818                    & 0.028                   & 0.041                    & 0.819                    & 0.034                   & 0.063                    \\ \hline
Median    & 0.827                    & 0.002                  & 0.017                    & 0.829                    & 0.033                   & 0.042                    & 0.819                    & 0.058                   & 0.084                    \\ \hline
TMean     & 0.831                    & 0.002                   & 0.014                    & 0.825                    & 0.040                    & 0.073                    & 0.820                     & 0.061                   & 0.089                    \\ \hline
FairFed & 0.831                    & 0.003                    & 0.020                    & 0.824                    & 0.040                   & 0.068                    & 0.823                    & 0.077                   & 0.126                     \\ \hline
FairTrade & 0.855                    & 0.010                    & 0.027                    & 0.841                    & 0.046                   & 0.060                    & 0.852                    & 0.049                   & 0.060                     \\ \hline
\end{tabular}
\end{table}

\begin{table}[t]
\caption{Fairness attack evaluation on Default dataset.}
\label{tab:default}
\centering
\scriptsize
%\tiny
\addtolength{\tabcolsep}{-2.66pt}
\begin{tabular}{|l|r|r|r|r|r|r|r|r|r|}
\hline
     & \multicolumn{3}{|c|}{No attack}& \multicolumn{3}{|c|}{One malicious client}& \multicolumn{3}{|c|}{Two malicious clients}\\ \hline
Scheme    & \multicolumn{1}{|l|}{Acc}    & \multicolumn{1}{|l|}{DP}    & \multicolumn{1}{|l|}{EOD} & \multicolumn{1}{|l|}{Acc}    & \multicolumn{1}{|l|}{DP}    & \multicolumn{1}{|l|}{EOD} & \multicolumn{1}{|l|}{Acc}    & \multicolumn{1}{|l|}{DP}    & \multicolumn{1}{|l|}{EOD} \\ \hline
FedAvg    & \multicolumn{1}{|r|}{0.622}   & \multicolumn{1}{|r|}{0.108} & 0.116                    & \multicolumn{1}{|r|}{0.630}    & \multicolumn{1}{|r|}{0.129} & 0.139                    & \multicolumn{1}{|r|}{0.649}   & \multicolumn{1}{|r|}{0.201} & 0.215                    \\ \hline
Krum      & \multicolumn{1}{|r|}{0.601}   & \multicolumn{1}{|r|}{0.023} & 0.018                    & \multicolumn{1}{|r|}{0.593}   & \multicolumn{1}{|r|}{0.047} & 0.048                    & \multicolumn{1}{|r|}{0.620}    & \multicolumn{1}{|r|}{0.056} & 0.055                    \\ \hline
MKrum     & \multicolumn{1}{|r|}{0.607}   & \multicolumn{1}{|r|}{0.107} & 0.109                    & \multicolumn{1}{|r|}{0.636}   & \multicolumn{1}{|r|}{0.144} & 0.156                    & \multicolumn{1}{|r|}{0.619}   & \multicolumn{1}{|r|}{0.146} & 0.157                    \\ \hline
Median    & \multicolumn{1}{|r|}{0.623}   & \multicolumn{1}{|r|}{0.091} & 0.094                    & \multicolumn{1}{|r|}{0.627}   & \multicolumn{1}{|r|}{0.125} & 0.131                    & \multicolumn{1}{|r|}{0.625}   & \multicolumn{1}{|r|}{0.160}  & 0.169                    \\ \hline
TMean     & \multicolumn{1}{|r|}{0.622}   & \multicolumn{1}{|r|}{0.110}  & 0.119                    & \multicolumn{1}{|r|}{0.626}   & \multicolumn{1}{|r|}{0.128} & 0.138                    & \multicolumn{1}{|r|}{0.639}   & \multicolumn{1}{|r|}{0.185} & 0.195                    \\ \hline
FairFed & \multicolumn{1}{|r|}{0.624}   & \multicolumn{1}{|r|}{0.106} & 0.114                    & \multicolumn{1}{|r|}{0.625}   & \multicolumn{1}{|r|}{0.125} & 0.136                    & \multicolumn{1}{|r|}{0.640}   & \multicolumn{1}{|r|}{0.213}  & 0.228                    \\ \hline
FairTrade & \multicolumn{1}{|r|}{0.662}   & \multicolumn{1}{|r|}{0.005} & 0.059                    & \multicolumn{1}{|r|}{0.691}   & \multicolumn{1}{|r|}{0.016} & 0.065                    & \multicolumn{1}{|r|}{0.649}   & \multicolumn{1}{|r|}{0.090}  & 0.109                    \\ \hline
\end{tabular}
\end{table}

\begin{table}[t]
\caption{Fairness attack evaluation on the Bank dataset.}
\label{tab:bank}
\centering
\scriptsize
%\tiny
\addtolength{\tabcolsep}{-2.66pt}
\begin{tabular}{|l|r|r|r|r|r|r|r|r|r|}
\hline
     & \multicolumn{3}{|c|}{No attack}& \multicolumn{3}{|c|}{One malicious client}& \multicolumn{3}{|c|}{Two malicious clients}\\ \hline
Scheme    & \multicolumn{1}{|l|}{Acc}    & \multicolumn{1}{|l|}{DP}    & \multicolumn{1}{|l|}{EOD} & \multicolumn{1}{|l|}{Acc}    & \multicolumn{1}{|l|}{DP}    & \multicolumn{1}{|l|}{EOD} & \multicolumn{1}{|l|}{Acc}    & \multicolumn{1}{|l|}{DP}    & \multicolumn{1}{|l|}{EOD} \\ \hline
FedAvg    & \multicolumn{1}{|r|}{0.830}    & \multicolumn{1}{|r|}{0.099} & 0.071                    & \multicolumn{1}{|r|}{0.829}   & \multicolumn{1}{|r|}{0.108} & 0.080                     & \multicolumn{1}{|r|}{0.828}   & \multicolumn{1}{|r|}{0.135} & 0.104                    \\ \hline
Krum      & \multicolumn{1}{|r|}{0.815}   & \multicolumn{1}{|r|}{0.102} & 0.073                    & \multicolumn{1}{|r|}{0.852}   & \multicolumn{1}{|r|}{0.106} & 0.088                    & \multicolumn{1}{|r|}{0.816}     & \multicolumn{1}{|r|}{0.110}  & 0.096                    \\ \hline
MKrum     & \multicolumn{1}{|r|}{0.824}   & \multicolumn{1}{|r|}{0.122} & 0.097                    & \multicolumn{1}{|r|}{0.828}   & \multicolumn{1}{|r|}{0.127} & 0.100                      & \multicolumn{1}{|r|}{0.825}   & \multicolumn{1}{|r|}{0.133} & 0.102                    \\ \hline
Median    & \multicolumn{1}{|r|}{0.830}    & \multicolumn{1}{|r|}{0.100}  & 0.080                     & \multicolumn{1}{|r|}{0.835}   & \multicolumn{1}{|r|}{0.109} & 0.082                    & \multicolumn{1}{|r|}{0.833}   & \multicolumn{1}{|r|}{0.117} & 0.086                    \\ \hline
TMean     & \multicolumn{1}{|r|}{0.832}   & \multicolumn{1}{|r|}{0.09}  & 0.066                    & \multicolumn{1}{|r|}{0.830}    & \multicolumn{1}{|r|}{0.112} & 0.084                    & \multicolumn{1}{|r|}{0.831}   & \multicolumn{1}{|r|}{0.121} & 0.091                    \\ \hline
FairFed & \multicolumn{1}{|r|}{0.830}    & \multicolumn{1}{|r|}{0.099} & 0.072                    & \multicolumn{1}{|r|}{0.830}   & \multicolumn{1}{|r|}{0.109} & 0.081                    & \multicolumn{1}{|r|}{0.827}   & \multicolumn{1}{|r|}{0.137} & 0.106                    \\ \hline
FairTrade & \multicolumn{1}{|r|}{0.860}    & \multicolumn{1}{|r|}{0.008} & 0.058                    & \multicolumn{1}{|r|}{0.866}   & \multicolumn{1}{|r|}{0.029} & 0.071                    & \multicolumn{1}{|r|}{0.863}   & \multicolumn{1}{|r|}{0.065} & 0.093                    \\ \hline
\end{tabular}
\end{table}

\begin{table}[!htbp]
\caption{Fairness attack evaluation on the KDD dataset.}
\label{tab:kdd}
\centering
\scriptsize
%\tiny
\addtolength{\tabcolsep}{-2.66pt}
\begin{tabular}{|l|r|r|r|r|r|r|r|r|r|}
\hline
    & \multicolumn{3}{|c|}{No attack}& \multicolumn{3}{|c|}{One malicious client}& \multicolumn{3}{|c|}{Two malicious clients}\\ \hline
Scheme    & \multicolumn{1}{|l|}{Acc}    & \multicolumn{1}{|l|}{DP}    & \multicolumn{1}{|l|}{EOD} & \multicolumn{1}{|l|}{Acc}    & \multicolumn{1}{|l|}{DP}    & \multicolumn{1}{|l|}{EOD} & \multicolumn{1}{|l|}{Acc}    & \multicolumn{1}{|l|}{DP}    & \multicolumn{1}{|l|}{EOD} \\ \hline
FedAvg    & \multicolumn{1}{|r|}{0.908}   & \multicolumn{1}{|r|}{0.150} & 0.316                    & \multicolumn{1}{|r|}{0.881}   & \multicolumn{1}{|r|}{0.186} & 0.326                    & \multicolumn{1}{|r|}{0.891}   & \multicolumn{1}{|r|}{0.193} & 0.374                    \\ \hline
Krum      & \multicolumn{1}{|r|}{0.901}   & \multicolumn{1}{|r|}{0.141} & 0.341                    & \multicolumn{1}{|r|}{0.913}   & \multicolumn{1}{|r|}{0.159} & 0.374                    & \multicolumn{1}{|r|}{0.892}   & \multicolumn{1}{|r|}{0.188} & 0.396                    \\ \hline
MKrum     & \multicolumn{1}{|r|}{0.903}   & \multicolumn{1}{|r|}{0.168} & 0.320                    & \multicolumn{1}{|r|}{0.907}   & \multicolumn{1}{|r|}{0.171} & 0.364                    & \multicolumn{1}{|r|}{0.912}   & \multicolumn{1}{|r|}{0.180} & 0.387                    \\ \hline
Median    & \multicolumn{1}{|r|}{0.909}   & \multicolumn{1}{|r|}{0.162} & 0.340                    & \multicolumn{1}{|r|}{0.898}   & \multicolumn{1}{|r|}{0.177} & 0.342                    & \multicolumn{1}{|r|}{0.890}   & \multicolumn{1}{|r|}{0.195} & 0.362                    \\ \hline
TMean     & \multicolumn{1}{|r|}{0.905}   & \multicolumn{1}{|r|}{0.166} & 0.329                    & \multicolumn{1}{|r|}{0.900}   & \multicolumn{1}{|r|}{0.176} & 0.347                    & \multicolumn{1}{|r|}{0.906}   & \multicolumn{1}{|r|}{0.181} & 0.380                    \\ \hline
FairFed & \multicolumn{1}{|r|}{0.912}   & \multicolumn{1}{|r|}{0.149} & 0.358                    & \multicolumn{1}{|r|}{0.886}   & \multicolumn{1}{|r|}{0.190} & 0.304                    & \multicolumn{1}{|r|}{0.891}   & \multicolumn{1}{|r|}{0.197} & 0.361                    \\ \hline
FairTrade & \multicolumn{1}{|r|}{0.883}   & \multicolumn{1}{|r|}{0.001} & 0.051                    & \multicolumn{1}{|r|}{0.891}   & \multicolumn{1}{|r|}{0.010} & 0.067                    & \multicolumn{1}{|r|}{0.867}   & \multicolumn{1}{|r|}{0.014} & 0.096                    \\ \hline
\end{tabular}
\end{table}

\subsection{Experimental Results and Discussion.}

\noindent\textbf{Attack evaluation.} Table~\ref{tab:adult} logs the result after executing the proposed attack utilizing DP-based fairness-constrained optimization. It can be observed that DP and EOD increase with all settings. With FedAvg, the DP increases 16.25\% and 25.50\% in the presence of one and two malicious participants, respectively. Byzantine-robust aggregation rules are also unable to detect the attack and mitigate bias in the global model. With Trimmed Mean, DP increases 9\% and 13.20\% in the presence of one and two malicious participants, respectively. However, they are able to restrict the impact with an increasing number of attackers. Applying fairness-aware aggregation like FairFed, DP increases 16.74\% and 26.10\% in the presence of one and two malicious participants, respectively. FairFed does not detect bias in the homogeneous data distribution system because its design is based on mitigating the bias based on imbalanced data. FairTrade is able to reduce the bias in a no-attack scenario. However, increasing the number of attackers from one to two causes an increase in DP by 59.18\% and 76.19\%, respectively. Although we have used DP-based fairness-constrained optimization for the attack, we can find a significant increase in EOD. It can also be observed that the attack maintains almost global accuracy. 

\noindent\textbf{Attribute-based data distribution.} In real-world FL environments, data is often partitioned according to inherent attributes rather than being distributed randomly between clients. For example, in a financial network for loan risk assessment, each bank or financial institution collects loan applicant data based on its local demographic and economic conditions, rather than a random sampling of applicants from across the region. To replicate such scenarios, we employ attribute-based data splitting among clients. Table~\ref{tab:adult-attr} provides the attack evaluation result with attribute-based data distribution. With FedAvg, the DP increases 46.86\% and 68.15\% in the presence of one and two malicious participants, respectively. With FairFed, DP increases by 46.44\% and 67. 03\% in the presence of one and two malicious participants, respectively. FairFed is more vulnerable to attack in this setting due to its design. Applying FairTrade, increasing the number of attackers from one to two causes an increase of DP by 32.83\% and 47.05\%. Similarly, we can observe a significant increase in EOD and almost maintained accuracy, even with a varying number of malicious participants. \textit{However, we can observe a noticeable decrease in accuracy in this case, due to the non-iid nature of the dataset.}
%\smallskip

%\noindent\textbf{Increasing number of participants.} We evaluate the attack with a larger number of participants, while keeping the number of malicious participants as one or two only. 
%Table~\ref{tab:adult-50} provides attack evaluation result with 50 participants. This shows the efficacy of the proposed attack in the presence of only 2\% and 4\% attackers. 
%With FedAvg, DP increases 2\% and 9.39\% in the presence of one and two malicious participants, respectively. Applying FairTrade, increasing the number of attackers from one to two causes an increase of DP by 53.33\% and 59.61\%.
%\smallskip

\noindent\textbf{Comparison.} Table~\ref{tab:comp1} provides the comparison for fairness attack on the Adult dataset with a poisoning attack DYN-OPT~\cite{shejwalkar2021manipulating} and a fairness attack EAB~\cite{ijcai2024p51}. It can be observed that our attack achieves more attack impact than EAB against all the aggregation rules in the presence of only one malicious participants out of a total of ten participants. DYN-OPT performs better in the case of Krum since it only selects one model for the next iteration and does not consider the one generated by our attack.
%\smallskip

\noindent\textbf{Attack Evaluation on other datasets.} Tables~\ref{tab:law}-\ref{tab:kdd} present the attack evaluation results over other datasets (Law, Default, Bank, and KDD) in the presence of a varying number of malicious participants under random data distribution settings. The attack impact against Law, Default and Bank datasets is significant both in terms of DP and EOD. KDD has an almost balanced distribution among the two groups of sensitive attributes and carries very little bias. With our attack on FedAvg, DP increases 19.35\% and 22.27\% in the presence of one and two malicious participants, respectively. Applying FairTrade, increasing the number of attackers from one to two, causes an increase of DP by 90\% and 92.85\%.

\section{Conclusion}
We presented a novel fairness attack targeting FL systems. Our attack exploits fairness constraints by embedding malicious objectives within the standard local optimization framework of FL. Through systematic manipulation of fairness metrics such as demographic parity and equal opportunity during local training, malicious clients degrade the fairness of the global model while maintaining global accuracy. This increased the bias up to 90\% and gained more attack impact than existing state-of-the-art poisoning and fairness attacks.
Our attack achieved a balance between stealthiness and impact, making it challenging for existing defense mechanisms to detect and mitigate.
%This work underscores the need for robust defense mechanisms capable of preserving fairness in FL while remaining resilient to adversarial manipulations.

\section*{Acknowledgment} 
This work is supported, in whole or in part, by the UKRI Prosperity Partnership Scheme (FAIR) under the EPSRC Grant EP/V056883/1; EP/R007195/1 (Academic Centre of Excellence in Cyber Security Research - University of Warwick); EP/N510129/1 (The Alan Turing Institute); the Bill and Melinda Gates Foundation [INV-001309]. Under the grant conditions of the Foundation, a Creative Commons Attribution 4.0 Generic License has already been assigned to the Author Accepted Manuscript version that might arise from this submission. The author gratefully acknowledges the support provided by the Department of Science and Technology (DST), Government of India, through the INSPIRE Faculty Fellowship scheme.

\bibliographystyle{plain}
\bibliography{tcom25}

\end{document}